\documentclass[conference]{IEEEtran}
\IEEEoverridecommandlockouts
\usepackage{cite}
\usepackage{amsmath,amssymb,amsfonts}
\usepackage{algorithmic}
\usepackage{graphicx}
\usepackage{textcomp}
\usepackage{xcolor}
\def\BibTeX{{\rm B\kern-.05em{\sc i\kern-.025em b}\kern-.08em
    T\kern-.1667em\lower.7ex\hbox{E}\kern-.125emX}}

\usepackage{float}
\usepackage{multirow}
\usepackage{hyperref}
\usepackage{url}
\usepackage{wrapfig}
\usepackage{cuted}
\usepackage{etoolbox}
\makeatletter
\patchcmd{\@makecaption}
  {\scshape}
  {}
  {}
  {}
\makeatother
\AfterEndEnvironment{strip}{\leavevmode}
\usepackage{capt-of}
\makeatletter
\newcommand{\ssymbol}[1]{^{\@fnsymbol{#1}}}
\makeatother
\definecolor{airforceblue}{rgb}{0.36, 0.54, 0.66}

\begin{document}

\title{CT-Bound: Robust Boundary Detection From Noisy Images Via Hybrid Convolution and Transformer Neural Networks}

\author{
\IEEEauthorblockN{Wei Xu\IEEEauthorrefmark{1}\thanks{* Corresponding author}, Junjie Luo, and Qi Guo}
\IEEEauthorblockA{\textit{Elmore Family School of Electrical and Computer Engineering} \\
\textit{Purdue University}, West Lafayette, IN, USA \\
\{\href{mailto:xu1639@purdue.edu}{xu1639},\href{mailto:luo330@purdue.edu}{luo330},\href{mailto:qiguo@purdue.edu}{qiguo}\}@purdue.edu}
\vspace{-0.4in}
}

\maketitle

\begin{strip}\centering
\vspace{-0.25in}
\includegraphics[width=1.0\textwidth]{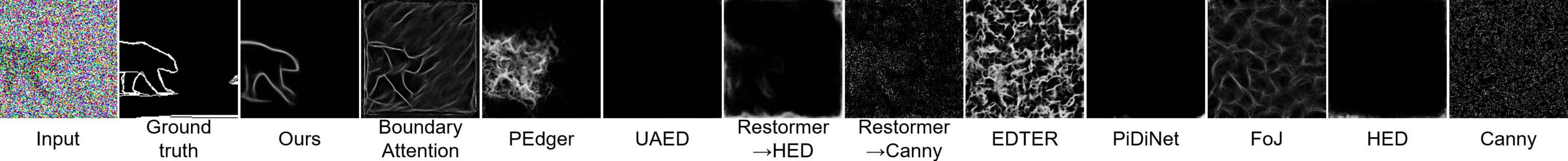}
\vspace{-0.1in}
\captionof{figure}{Boundary detection from noisy images. Compared to a variety of models~\cite{mia2023boundaries,fu2023practical,zhou2023treasure,zamir2022restormer,xie2015holistically,canny1986computational,pu2022edter,su2021pixel,verbin2021field}, ours robustly detects the boundaries even when they are visually challenging to discriminate.}
\label{fig:teaser}
\vspace{-0.15in}
\end{strip}

\begin{abstract}
We present CT-Bound, a robust and fast boundary detection method for very noisy images using a hybrid Convolution and Transformer neural network. The proposed architecture decomposes boundary estimation into two tasks: local detection and global regularization. During the local detection, the model uses a convolutional architecture to predict the boundary structure of each image patch in the form of a pre-defined local boundary representation, the field-of-junctions (FoJ)~\cite{verbin2021field}. Then, it uses a feed-forward transformer architecture to globally refine the boundary structures of each patch to generate an edge map and a smoothed color map simultaneously. Our quantitative analysis shows that CT-Bound outperforms the previous best algorithms in edge detection on very noisy images. It also increases the edge detection accuracy of FoJ-based methods while having a 3-time speed improvement. Finally, we demonstrate that CT-Bound can produce boundary and color maps on real captured images without extra fine-tuning and real-time boundary map and color map videos at ten frames per second.
\end{abstract}


\vspace{-0.05in}
\section{Introduction}
\label{sec:intro}

Detecting boundary structures from very noisy images is a common and challenging computer vision problem~\cite{sun2022survey}. There have been many applications that require boundary detection from images with very low light levels, such as medical imaging, manufacturing, autonomous navigation, etc. Although image boundary detection has been broadly studied since the early stage of computer vision~\cite{irwin2014isotropic, arbelaez2010contour, malik2001contour, canny1986computational, roberts1965machine}, our results show that the accuracy of current best boundary detection algorithms are still unsatisfactory when the input images have a very low light level (Fig.~\ref{fig:teaser}). 

We present CT-Bound. It is a deep neural network architecture that can robustly detect boundaries from a single noisy image. The model processes the input image to predict a generalized local boundary representation for each image patch called the field-of-junctions (FoJs)~\cite{verbin2021field}. FoJ can represent a variety of boundary types in image patches, including edges, corners, and contours, and is an effective prior in edge detection, especially for noisy images~\cite{verbin2021field, mia2023boundaries}. By constraining the predicted boundary structures to those that FoJ can describe, we observe our model can detect very faint edge signals in the presence of significant noise (Fig.~\ref{fig:teaser}). Our experiment shows that CT-Bound achieves the highest boundary detection accuracy among a variety of recent edge detection methods. 

CT-Bound consists of an innovative two-stage, hybrid Convolution and Transformer neural network architecture (Fig.~\ref{fig:model}). The first stage is a convolutional architecture that makes an initial prediction of a local FoJ parameterization solely based on the visual appearance of each image patch. The second stage consists of a feedforward transformer encoder that takes in the initial FoJ estimation of all image patches to perform refinement. The architecture is novel as it completely decomposes boundary estimation into two tasks: detecting boundaries from local image patch and regularizing neighboring boundary estimations to ensure consistency and to look like natural boundaries. The convolutional network stage conducts boundary detection using a small receptive field ($21 \times 21$ in our experiments). Thus, it does not need to learn the global appearance of images and can be trained using synthetic image patches that only contain basic boundary structures. The transformer stage only receives the FoJ representation and has no access to the input image during inference. Therefore, the computational complexity of the transformer in the CT-Bound is significantly lower than that of the classic Vision Transformer~\cite{dosovitskiy2020image}.

Besides verifying the accuracy and robustness of the proposed algorithm using noisy images simulated from standard, benchmark datasets under different noise levels, we also test CT-Bound using noisy images captured using real-world cameras to generate real-time videos of boundary maps. The contribution of this paper can be summarized as follows. 
\begin{itemize}
    \item A two-stage, hybrid neural network architecture;
    \item A robust and fast and non-iterative solver of FoJ that enables real-time boundary detection on very noisy images;
    \item A thorough experimental study that demonstrates CT-Bound achieves the highest or among the highest accuracy in detecting image boundaries from very noisy images compared to the previous best algorithms.
\end{itemize}
The code, the training data, the testing data, additional results, and the video demonstration of the proposed method can be found at \href{https://github.com/guo-research-group/CT-Bound}{\textcolor{blue}{https://github.com/guo-research-group/CT-Bound}}.

\vspace{-0.05in}
\section{Related Work}
\label{sec:related}

According to the classification by Gong et al.~\cite{gong2018overview}, image boundary detection methods can be divided into four categories, i.e., boundaries from luminance changes, texture changes, perceptual grouping, and illusory contour. We focus our literature review on the first category to which the proposed method belongs.

\paragraph{Local boundary detection} The first step of boundary detection is typically using specially designed filters to locate local responses of image boundaries. The filters either maximize the detectability and localization accuracy of the boundaries under noise, such as the Roberts cross operator~\cite{roberts1965machine}, the Canny detectors~\cite{canny1986computational}, the Laplacian detectors~\cite{wang2007laplacian}, and the perfect matching filters~\cite{ofir2019detection} or are sensitive to the direction of the image boundaries, e.g., Sobel filters~\cite{irwin2014isotropic}, Gaussian quadrature pairs~\cite{malik2001contour}, Steerable filters~\cite{freeman1991design}, etc. To robustly detect boundaries under non-ideal edges, people also develop sophisticated filters that are non-linear~\cite{perona1991detecting} or operate in multiple scales~\cite{lu1992reasoning}. However, these classic methods based on local patches are found to be insufficient when the noise of the image is severe and have limited visual information to confidently determine the edges from the receptive field of a single filter. In this case, image smoothing is also not a solution as the smoothing operation will make faint or fine boundary structures to be indistinguishable~\cite{ofir2019detection}.

\paragraph{Global boundary refinement} Boundaries in natural images are usually piece-wise smooth. Based on this observation, people have developed algorithms to refine the global boundary map from the locally detected boundaries by regularizing the curvature, for example, the squared curvature~\cite{nieuwenhuis2014efficient} or total curvature~\cite{zhong2020minimizing} along the boundary. Another observation for global boundary refinement is that the neighboring boundary maps must agree. There have been methods that use this intuition by enforcing neighboring consistency~\cite{verbin2021field, ofir2016fast, arbelaez2010contour}. These global refinement methods are typically iterative and thus could have a high computational complexity. 

\paragraph{Deep learning boundary detection} The emergence of deep learning has enabled people to develop deep neural networks that fuse the two steps into an end-to-end architecture learned from data. These methods directly output global boundary maps in nonparametric ways~\cite{zhou2023treasure, fu2023practical, pu2022edter, su2021pixel, xie2015holistically, poma2020dense, kirillov2023segment} or parametric ways~\cite{huang2018learning, xue2019learning}. These methods typically outperform traditional, non-learning-based edge detection methods nowadays. There is a recent work, Boundary Attention, that also combines FoJ representation with deep neural networks~\cite{mia2023boundaries}. It is targeted for detecting complicated, fine boundary structures and outputting more in-depth boundary information, such as edge-aware distance maps, from noisy images. Compared with Boundary Attention, CT-Bound focuses on boundary detection and achieves a higher boundary detection accuracy on benchmark datasets with a 3-time faster speed. Nonetheless, we suggest readers also read the paper~\cite{mia2023boundaries} for a more comprehensive understanding of this field. 

\vspace{-0.05in}
\section{Methods}
\label{sec:method}

\begin{wrapfigure}{R}{2.8cm}
\vspace{-0.15in}
\centering
\includegraphics[width=2.5cm]{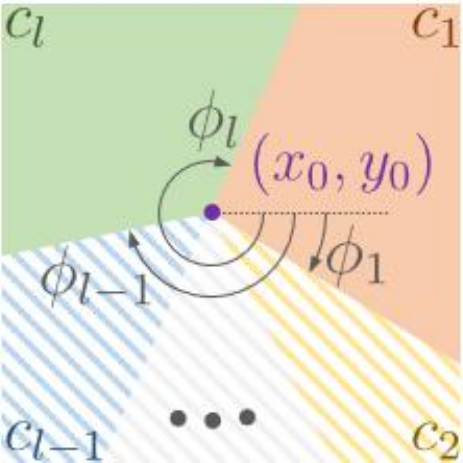}
\vspace{-0.05in}
\caption{Field-of-junction (FoJ) representation~\cite{verbin2021field}.}
\label{fig:foj_rep}
\vspace{-0.1in}
\end{wrapfigure}

We first briefly describe the FoJ representation~\cite{verbin2021field} that we adopt in this work. Given an image patch $P\in \mathbb{R}^{h\times w \times k}$ with dimension $h\times w$ and $k$ color channel, the FoJ models its
boundary structure using a parameter set $\Phi = (\boldsymbol{x}, \boldsymbol{\phi}, \boldsymbol{c})$, where $\boldsymbol{x} = (x_0, y_0)$ indicates the center of the vertex, $\boldsymbol{\phi} = (\phi_1, \cdots, \phi_l)$ represents the angles of the $l$ edges, $\boldsymbol{c} = (\boldsymbol{c}_1, \cdots, \boldsymbol{c}_l), \boldsymbol{c}_j \in \mathbb{R}^k, j=1,\cdots, l$ are the color of the region between every pair of neighboring edges. The parameter $l$ is a hyperparameter that needs to be predetermined. See Fig.~\ref{fig:foj_rep} for an exemplar illustration of FoJ. As shown in Verbin and Zickler, FoJ can represent a variety of local boundary structures, including edges, corners, and junctions~\cite{verbin2021field}. Given an FoJ representation $\Phi$, the corresponding boundary map of the patch $B(x,y;\Phi)$ can be plotted via:
\vspace{-0.01in}
\begin{align}
    B(x,y;\Phi) = \pi \epsilon H_{2,\epsilon}^{\prime}(\min(d_j(x,y))),
    \label{eq:patch-boundary}
\end{align}
where $d_j(x,y)$ is the distance from the pixel $(x,y)$ to the edge $j$ and $H_{2,\epsilon}^{\prime}$ is the derivative of Heaviside function $H_{2,\epsilon}$ in~\cite{chan2001active}:
\vspace{-0.01in}
\begin{align}
    H_{2,\epsilon}(d) = \frac{1}{2} \left( 1+\frac{2}{\pi} \arctan \frac{d}{\epsilon}\right),
\end{align}
where $\epsilon$ is a smoothing parameter that we set $\epsilon = 0.01$ throughout our experiments. The color map $C(x,y;\Phi)$ can be visualized using:
\vspace{-0.01in}
\begin{align}
    C(x,y;\Phi) = \sum_{j=1}^l \delta_j(x,y) \boldsymbol{c}_j,
    \label{eq:patch-color}
\end{align}
where $\delta_j(x,y) = 1$ when the pixel $(x,y)$ is within the wedge between edge $j$ and $j+1$ in the patch $P$:
\vspace{-0.01in}
\begin{equation}
\begin{aligned}
    \Omega_{j} = \{&(x,y) | (x,y) \in P, \\
    &(x - x_0) \cos \phi_j + (y - y_0) \sin\phi_j < 0, \\
    &(x - x_0) \cos \phi_{j+1} + (y - y_0) \sin\phi_{j+1} > 0
    \},
    \label{eq:wedge}
\end{aligned}
\end{equation}
and $\delta_j(x,y) = 0$ otherwise. 

\vspace{-0.05in}
\subsection{Network architecture}
\label{sec:method:rep}

\begin{figure*}[htb]
\centering
\centerline{\includegraphics[width=15cm]{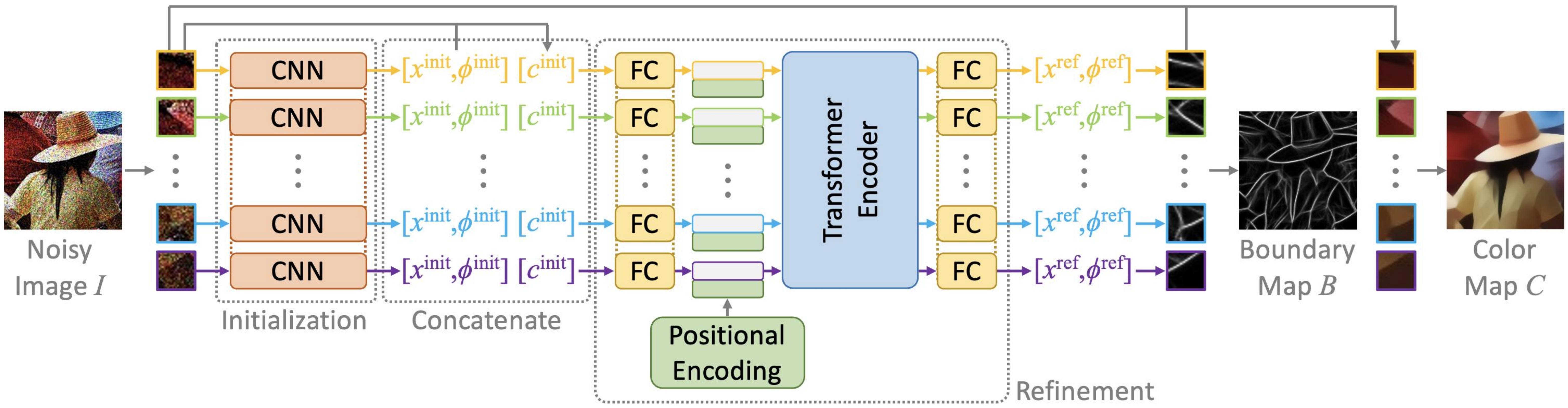}}
\vspace{-0.05in}
\caption{Network architecture of CT-Bound. The architecture consists of two stages. The initialization stage contains shared-weights convolutional neural networks that output the FoJ representation of every image patch. The refinement stage contains a transformer encoder that simultaneously refines all per-patch FoJ representations. Finally, the framework combines all per-patch FoJ representations together to output the global boundary map and the color map.}
\label{fig:model}
\vspace{-0.2in}
\end{figure*}

The network architecture of the proposed method is visualized in Fig.~\ref{fig:model}. Given a noisy image $I\in \mathbb{R}^{H\times W \times k}$, CT-Bound first divides the image $I$ into overlapping patches. The initialization stage takes in each image patch $P_{m,n}$ into a CNN to generate the initial vertex location and the edge angles of the FoJ representation $(\boldsymbol{x}^{\text{init}}_{m,n}, \boldsymbol{\phi}^{\text{init}}_{m,n})$. Then, the method determines the color parameters $\boldsymbol{c}^{\text{init}}_{m,n} = (\boldsymbol{c}^{\text{init}}_{m,n,1}, \cdots, \boldsymbol{c}^{\text{init}}_{m,n,l})$ mathematically by averaging the color of pixels of each divided area of the patch:
\vspace{-0.01in}
\begin{align}
    \boldsymbol{c}^{\text{init}}_{m,n,j} = \frac{1}{|\Omega_{m,n,j}|}\sum_{(x,y)\in \Omega_{m,n,j}} P_{m,n}(x,y,:),
    \label{eq:color}
\end{align}
where $\Omega_{m,n,j}$ indicates the set of pixels in the wedge between edges $j$ and $j+1$ in the patch $P_{m,n}$, as defined in \eqref{eq:wedge}.

The refinement stage takes in the initial FoJ representation $(\boldsymbol{x}^{\text{init}}_{m,n}, $ $\boldsymbol{\phi}^{\text{init}}_{m,n},$ $\boldsymbol{c}^{\text{init}}_{m,n})$ of all patches $P_{m,n}$ simultaneously. It first converts each initial FoJ representation $(\boldsymbol{x}^{\text{init}}_{m,n}, \boldsymbol{\phi}^{\text{init}}_{m,n},$ $\boldsymbol{c}^{\text{init}}_{m,n})$ into a feature vector representation $\boldsymbol{v}_{m,n}\in \mathbb{R}^d$, and applies positional encoding by adding a positional vector $\boldsymbol{p}_{m,n} = [p_{m,n,1}, \cdots, p_{m,n,d}]^T$ to the feature vector $\boldsymbol{v}_{m,n}$ to incorporate the positional information of each image patch. The 2D positional encoding vector follows the design of Zhang and Liu~\cite{wang2021translating}:
\vspace{-0.01in}
\begin{align*}
    p_{m,n,i} &= \begin{cases}
        \sin \left( \frac{m}{10000^{4i/D}} \right), i = 0, 2, 4, \cdots, d/2 \\
        \cos \left( \frac{m}{10000^{4i/D}} \right), i = 1, 3, 5, \cdots, d/2 + 1\\
        \sin \left( \frac{n}{10000^{4i/D}} \right), i = d/2, d/2+2, \cdots, d - 2 \\
        \cos \left( \frac{n}{10000^{4i/D}} \right), i = d/2+1, d/2+3, \cdots, d - 1
    \end{cases}
\end{align*}
where the dimension of each feature vector $d$ is an even number. All positional encoded feature vectors are fed into a transformer encoder consisting of a series of multi-head attention layers to refine the boundary consistency among patches globally and adjust unnatural boundary estimations. It is the only block in the framework that globally shares the per-patch FoJ information. Finally, the framework outputs the refined vertex location and edge angles of all patches $(\boldsymbol{x}^{\text{ref}}_{m,n}, \boldsymbol{\phi}^{\text{ref}}_{m,n}), \forall m,n$, and calculate the refined color parameters $\boldsymbol{c}^{\text{ref}}_{m,n}$ using \eqref{eq:color}. We list the network hyperparameters we use in our experiment in Tab.~\ref{tab:cnn}. As the transformer does not operate on the image domain, the dimension of the input vector and the number of layers are much smaller compared to the classic Vision Transformer~\cite{dosovitskiy2020image}.

\begin{table}[tb]
\caption{CT-Bound model hyperparameter.}
\vspace{-0.08in}
\label{tab:cnn}
\centering
\begin{tabular}{ccc}
\hline
\multicolumn{3}{c}{\textcolor{orange}{Convolutional Neural Network}} \\
Layer & Specification & Output \\
\hline
Conv2d & 5$\times$5 kernel, 4 stride, 2 pad & (21, 21, 96) \\
MaxPool2d & 3$\times$3 kernel, 2 stride, 0 pad & (10, 10, 96) \\
Conv2d & 5$\times$5 kernel, 1 stride, 2 pad & (10, 10, 256) \\
MaxPool2d & 2$\times$2 kernel, 2 stride, 0 pad & (5, 5, 256) \\
Conv2d & 3$\times$3 kernel, 1 stride, 1 pad & (5, 5, 384) \\
Conv2d & 3$\times$3 kernel, 1 stride, 1 pad & (5, 5, 384) \\
Conv2d & 3$\times$3 kernel, 1 stride, 1 pad & (5, 5, 256) \\
MaxPool2d & 3$\times$3 kernel, 2 stride, 0 pad & (2, 2, 256) \\
FC & - & 4096 \\
FC & - & 1024 \\
FC & - & 5 \\
\hline
\hline
\multicolumn{3}{c}{\textcolor{airforceblue}{Transformer Encoder}} \\
\multicolumn{2}{c}{Specification} & Parameter \\
\hline
\multicolumn{2}{c}{Dimension of each input vector} & 128 \\
\multicolumn{2}{c}{Number of layers} & 8 \\
\multicolumn{2}{c}{Number of heads in each layer} & 8 \\
\multicolumn{2}{c}{Dimension of the feed-forward layer} & 256 \\
\hline
\end{tabular}
\vspace{-0.15in}
\end{table}

From the refined FoJ parameters $\Phi^{\text{ref}}_{m,n} = (\boldsymbol{x}^{\text{ref}}_{m,n}, $ $ \boldsymbol{\phi}^{\text{ref}}_{m,n},$ $ \boldsymbol{c}^{\text{ref}}_{m,n})$, CT-Bound generates the per-patch boundary and color map according to \eqref{eq:patch-boundary} and \eqref{eq:patch-color}, and computes the global boundary map $B(x,y)$ by averaging the per-patch boundary maps~\cite{verbin2021field}:
\vspace{-0.01in}
\begin{align}
    B(x,y) = \frac{1}{N(x,y)} \sum_{N(x,y)} B(x,y; \Phi^{\text{ref}}_{m,n}),
\end{align}
and the global color map $C(x,y)$ via a specific smoothing operation over the per-patch color maps:
\vspace{-0.01in}
\begin{align}
    C(x,y) = \frac{1}{\left|N(x,y)\right|} \sum_{N(x,y)} \delta_{m,n,j}(x,y) \boldsymbol{c}^{\text{ref}}_{m,n,j},
\end{align}
where $N(x,y) = \left\{(m,n)|(x,y) \in \Omega_{m,n,j} \right\}$ is the set of the patch indices that contain $(x,y)$, $\delta_{m,n,j}$ is a binary indicator that is $1$ if pixel $(x,y)$ belongs to the wedge $\Omega_{m,n,j}$ and $0$ otherwise, and $\boldsymbol{c}^{\text{ref}}_{m,n,j}$ is the refined color of wedge $\Omega_{m,n,j}$. 

\vspace{-0.05in}
\subsection{Loss functions}
\label{sec:method:loss}

We develop a multi-stage training scheme to optimize the parameters of CT-Bound. First, we train the initialization stage using the patch reconstruction loss:
\vspace{-0.01in}
\begin{align}
    \mathcal{L}_{\text{init}}  = \mathbb{E}_{P} \left( \mathrm{MSE}\left(C\left(x,y;\Phi^{\text{gt}}\right), C\left(x,y;\Phi^{\text{init}}\right) \right) \right),
    \label{eq:cnn_loss}
\end{align}
where $\mathbb{E}_P$ denotes the expectation over all patches in the training set, and $C(x,y;$ $\Phi^{\text{gt}})$ and  $C(x,y;\Phi^{\text{init}})$ indicate the per-patch color maps reconstructed using true and estimated FoJ parameters, respectively. We observe that the visual quality of the FoJ estimation is higher when using the loss in \eqref{eq:cnn_loss} for training than directly supervising the FoJ parameters. Furthermore, because the CNN in the initialization stage has a small receptive field, we can use synthetic image patches of basic shapes to train it, and we observe that the trained model can be generalized to real-world image patches without further fine-tuning.

When optimizing parameters of the refinement stage, we use a fixed, pre-trained initialization stage to generate inputs $\Phi^{\text{init}}$. We invent a two-step training process for optimizing the refinement stage, which was noticed to lead to a more stable and faster convergence. In the first step, a mean squared error loss function is adopted to supervise the estimated FoJ parameters directly: 
\vspace{-0.01in}
\begin{align}
    \mathcal{L}_{\text{ref1}}  = \mathbb{E}_P\left( \lVert \boldsymbol{x}^{gt} - \boldsymbol{x}^{\text{ref}} \rVert^2 +  \lVert \boldsymbol{\phi}^{gt} - \boldsymbol{\phi}^{\text{ref}} \rVert^2 \right).
    \label{eq:step1_loss}
\end{align}
In the second step, we use a comprehensive image reconstruction loss adapted from Verbin and Zickler~\cite{verbin2021field}:
\vspace{-0.01in}
\begin{align}
    \mathcal{L}_{\text{ref2}} = \mathbb{E}_I\left(l_p  + \lambda_b l_b + \lambda_c l_c \right),
    \label{eq:final_loss}
\end{align}
where $\mathbb{E}_I$ indicates the expectation over all images in the dataset, and $l_p$, $l_b$, and $l_c$ are patch, boundary, and color loss terms, respectively:
\vspace{-0.01in}
\begin{align*}
    l_p &= \sum_{m,n} \sum_{j=1}^l \sum_{x,y} \delta_{m,n,j}(x,y) \lVert \boldsymbol{c}^{\text{ref}}_{m,n,j} - I(x,y) \rVert^2, \\
    l_b &= \sum_{m,n} \sum_{x,y} \left(B(x,y) - B(x,y;\Phi^{\text{ref}}_{m,n}) \right)^2, \\
    l_c &= \sum_{m,n} \sum_{j=1}^l \sum_{x,y} \delta_{m,n,j}(x,y) \lVert  \boldsymbol{c}^{\text{ref}}_{m,n,j} -  C(x,y) \rVert^2.
\end{align*}
In Verbin and Zickler, the loss in \eqref{eq:final_loss} was solved in an alternating, two-step fashion to refine the FoJ representation iteratively~\cite{verbin2021field}. We evaluate the loss in a single step and show it can successfully fine-tune the feed-forward transformer encoder to improve the FoJ representation.

\vspace{-0.05in}
\section{Experimental Results}
\label{sec:experiment}

\vspace{-0.05in}
\subsection{Data processing}
\label{sec:experiment:dataset}

For the training of the initialization stage, we use randomly sampled patches from FoJ synthetic datasets~\cite{verbin2021field} that only contain images of basic shapes such as squares. We select $8000$ image patches for training and $2000$ for testing. To simulate image noise, we apply a Poisson-Gaussian process to each image patch~\cite{ding2016modeling}:
\vspace{-0.01in}
\begin{align}
    P(x,y) = \mathrm{Poisson}\left(\alpha P^*(x,y)\right) + \mathrm{Gaussian}\left(0,\sigma^2\right),
    \label{eq:noise}
\end{align}
where $P(x,y)$ and $P^*(x,y) \in [0,1]$ are the noisy and normalized clean image patches, $\alpha$ is the photon level parameter that controls the noise of the image, and $\sigma$ is the standard deviation of read noise ($\sigma = 2$ is applied as in~\cite{chan2022does}). For the refinement stage, we use images from MS COCO~\cite{lin2014microsoft} for training and testing. The training and testing sets contain $1600$ and $400$ randomly selected, non-overlapping images, respectively. Each image is cropped at the center to the size of $147 \times 147$ and is applied with the same noise as described in \eqref{eq:noise}. In our experiment, we randomly set the photon level $\alpha$ within the range $[2,10]$ to generate images with a variety of noise levels. 

\begin{figure}[tb]
\centering
\includegraphics[width=0.9\columnwidth]{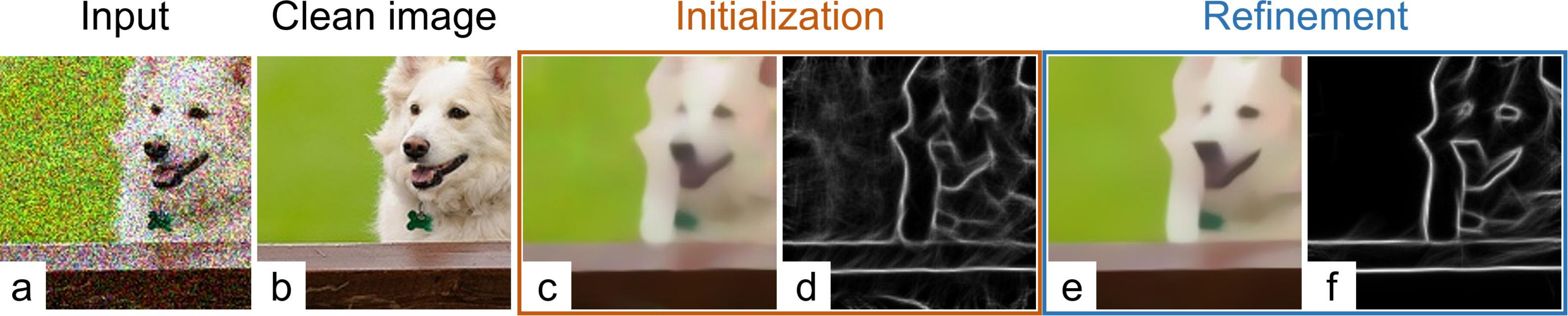}
\vspace{-0.1in}
\caption{Effect of refinement and boundary selection. (a-b) An input noisy image and the corresponding clean image from the MS COCO dataset~\cite{lin2014microsoft}. (c-d) The color map and boundary map before the refinement stage. (e-f) The color map and boundary map after the refinement stage. Noisy edge estimations are removed in refinement,
and the color map is smoother.}
\label{fig:cnn_trans_ex}
\vspace{-0.13in}
\end{figure}

\begin{table}[t]
\caption{ODS F1-score of CT-Bound before (numbers in \textcolor{orange}{red}) and after (numbers in \textcolor{airforceblue}{blue}) the refinement.}
\vspace{-0.1in}
\label{tab:diff_cnn}
\centering
\setlength{\tabcolsep}{1.7mm}{
\begin{tabular}{c|cc|cc}
\hline
\multirow{2}{*}{\shortstack{Photon level\\$\alpha_{\text{test}}$}} & \multicolumn{4}{c}{Dataset} \\
\cline{2-5}
& \multicolumn{2}{c|}{BSDS500~\cite{arbelaez2010contour}} & \multicolumn{2}{|c}{NYUDv2~\cite{silberman2012indoor}} \\
\hline
2 & \textcolor{orange}{0.482} & \textcolor{airforceblue}{0.541} & \textcolor{orange}{0.479} & \textcolor{airforceblue}{0.552} \\
4 & \textcolor{orange}{0.509} & \textcolor{airforceblue}{0.627} & \textcolor{orange}{0.522} & \textcolor{airforceblue}{0.633} \\
6 & \textcolor{orange}{0.518} & \textcolor{airforceblue}{0.640} & \textcolor{orange}{0.538} & \textcolor{airforceblue}{0.646} \\
8 & \textcolor{orange}{0.524} & \textcolor{airforceblue}{0.633} & \textcolor{orange}{0.546} & \textcolor{airforceblue}{0.647} \\
\hline
\end{tabular}}
\vspace{-0.2in}
\end{table}

We evaluate CT-Bound on the testing sets of Berkeley Segmentation Data Set 500 (BSDS500)~\cite{arbelaez2010contour} and NYU Depth Dataset V2 (NYUDv2)~\cite{silberman2012indoor}. We crop images to $147 \times 147$ size and add noise as above. BSDS500 has 200 testing images. For NYUDv2, 200 images are randomly selected from its testing set split and adopted by in~\cite{gupta2013perceptual}. Using different datasets for evaluation demonstrates the generalizability of our model. 

\vspace{-0.05in}
\subsection{Implementation details}
\label{sec:experiment:implement}

We use $l=3$ in our implementation. All optimizations in this work use the Adam optimizer~\cite{kingma2014adam}. The initialization stage is trained with an initial learning rate of $0.0002$ and a decay of $0.5$ every $80$ epochs. The batch size is $32$, and the total number of training epochs is $900$. We use a two-step scheme to train the refinement stage as described in Sec.~\ref{sec:method:loss}. Both steps use a batch size $16$. The first step uses \eqref{eq:step1_loss} as the objective function and has $100$ epochs. a learning rate $5 \times 10^{-5}$. The second step switches to \eqref{eq:final_loss} as its loss function and runs $1600$ epochs. The learning rate for the second step is updated with a triangular cycle between $1.75 \times 10^{-4}$ and $3.5 \times 10^{-4}$. The training and testing are performed on a machine with an NVIDIA GeForce RTX A5000 graphics card and 24 GB memory. 

The fixed contour threshold (ODS) F1-score is recorded during evaluation, with a non-maximum suppression~\cite{canny1986computational} applied in advance. We adjust the localization tolerance proportionally based on~\cite{xie2015holistically} to accommodate the image size in our experiment, setting it to 0.0209 for BSDS500 and 0.0372 for NYUDv2. 

\vspace{-0.05in}
\subsection{Ablation study}
\label{sec:experiment:ablation}

\begin{table*}[tb]
\vspace{-0.01in}
\caption{Quantitative comparison of boundary detection on noisy images synthesized from BSDS500~\cite{arbelaez2010contour} and NYUDv2~\cite{silberman2012indoor} datasets. The numbers are ODS F1-score. The proposed method and Boundary Attention~\cite{mia2023boundaries} demonstrate the highest or the second highest boundary detection accuracy on both datasets when the image is very noisy, i.e., $\alpha_{\text{test}}=2$. Meanwhile, ours is three times faster than Boundary Attention~\cite{mia2023boundaries}. When the noise level becomes lower, i.e., $\alpha_{\text{test}}$ being larger, Restormer~\cite{zamir2022restormer}$\rightarrow$HED~\cite{xie2015holistically} starts to perform better. This result suggests the robustness of our method in detecting boundaries at extremely high noise levels.}
\vspace{-0.05in}
\label{tab:acuracy}
\centering
\begin{tabular}{c|c|cccc|cccc|c}
\hline
\multirow{3}{*}{Model} & \multirow{3}{*}{Publication'Year} & \multicolumn{4}{|c|}{BSDS500~\cite{arbelaez2010contour}} & \multicolumn{4}{|c|}{NYUDv2~\cite{silberman2012indoor}} & FPS \\
\cline{3-10}
& & \multicolumn{8}{|c|}{Photon level $\alpha_{\text{test}}$} & \scriptsize{$\ssymbol{4}$OpenCV-CPU} \\
\cline{3-10}
& & 2 & 4 & 6 & 8 & 2 & 4 & 6 & 8 & \scriptsize{$\ssymbol{2}$PyTorch-GPU $\ssymbol{3}$JAX-GPU} \\
\hline
Canny~\cite{canny1986computational} & PAMI'86 & 0.493 & 0.489 & 0.383 & 0.495 & 0.476 & 0.467 & 0.325 & 0.482 & 625$\ssymbol{4}$ \\
HED~\cite{xie2015holistically} & ICCV'15 & 0.327 & 0.388 & 0.456 & 0.520 & 0.307 & 0.363 & 0.434 & 0.484 & 18.8$\ssymbol{4}$ \\
FoJ~\cite{verbin2021field} & ICCV'21 & 0.509 & 0.564 & 0.597 & 0.611 & 0.529 & 0.576 & 0.597 & 0.613 & 1/68$\ssymbol{2}$ \\
PiDiNet~\cite{su2021pixel} & ICCV'21 & 0.316 & 0.356 & 0.447 & 0.480 & 0.274 & 0.311 & 0.409 & 0.456 & 125$\ssymbol{2}$ \\
EDTER~\cite{pu2022edter} & CVPR'22 & 0.484 & 0.509 & 0.545 & 0.594 & 0.488 & 0.478 & 0.534 & 0.581 & 15.6$\ssymbol{2}$ \\
Restormer~\cite{zamir2022restormer}$\rightarrow$Canny~\cite{canny1986computational} & CVPR'22+PAMI'86 & 0.490 & 0.521 & 0.518 & 0.516 & 0.511 & 0.533 & 0.503 & 0.499 & 9.6$\ssymbol{2}$ \\
Restormer~\cite{zamir2022restormer}$\rightarrow$HED~\cite{xie2015holistically} & CVPR'22+ICCV'15 & 0.459 & \underline{\textbf{0.628}} & \underline{\textbf{0.674}} & \underline{\textbf{0.707}} & 0.474 & \textbf{0.625} & \underline{\textbf{0.647}} & \underline{\textbf{0.663}} & 6.4$\ssymbol{2}$ \\
UAED~\cite{zhou2023treasure} & CVPR'23 & 0.473 & 0.553 & 0.618 & 0.665 & 0.452 & 0.557 & 0.620 & \textbf{0.652} & 28.7$\ssymbol{2}$ \\
PEdger~\cite{fu2023practical} & ACMMM'23 & 0.525 & 0.611 & \textbf{0.657} & \textbf{0.684} & 0.509 & 0.606 & 0.632 & 0.650 & 14.4$\ssymbol{2}$ \\
Boundary Attention~\cite{mia2023boundaries} & arXiv'24 & \textbf{0.534} & 0.591 & 0.607 & 0.615 & \underline{\textbf{0.572}} & 0.609 & 0.624 & 0.625 & 4.3$\ssymbol{3}$ \\
\hline
\textbf{Ours} & - & \underline{\textbf{0.541}} & \textbf{0.627} & 0.640 & 0.633 & \textbf{0.552} & \underline{\textbf{0.633}} & \textbf{0.646} & 0.647 & 11.4$\ssymbol{2}$, 15.2$\ssymbol{3}$ \\
\hline
\end{tabular}
\end{table*}

\begin{figure*}[h]
\centering
\includegraphics[width=1.0\textwidth]{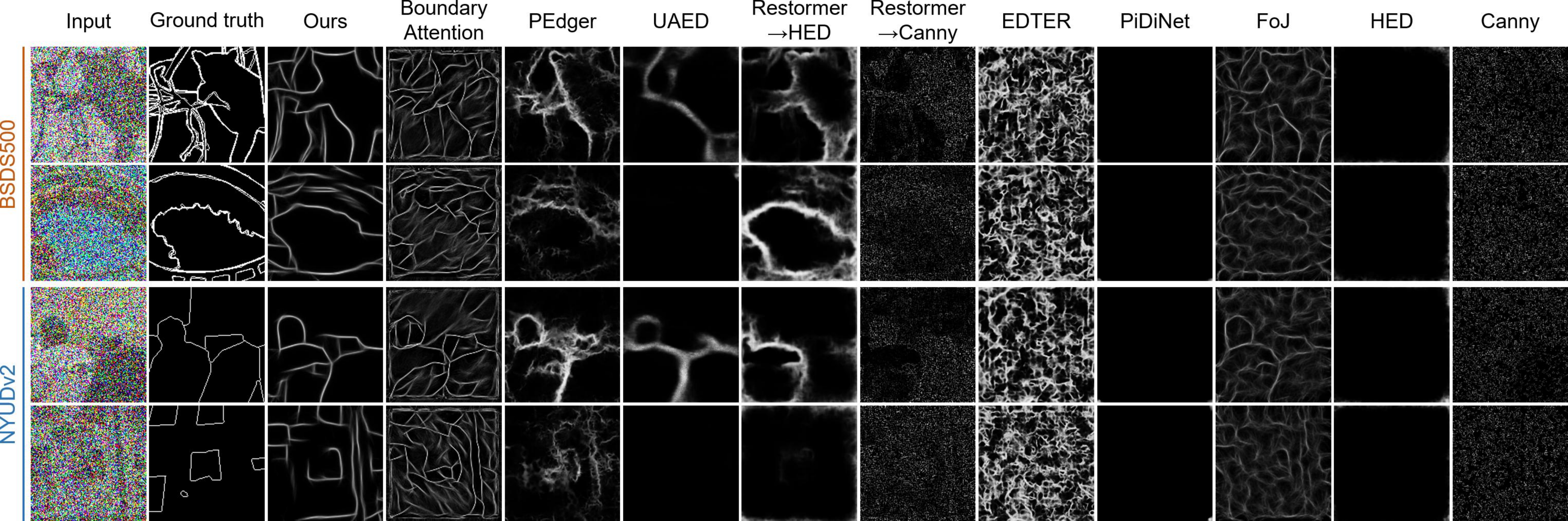}
\caption{Qualitative comparison of noisy images synthesized from BSDS500~\cite{arbelaez2010contour} and NYUDv2~\cite{silberman2012indoor} datasets with \textcolor{red}{photon level $\alpha_{\text{test}}=2$}. The proposed method shows robustness to the high noise level, while other methods fail to produce accurate boundaries. Additionally, ours can detect faint boundaries that are visually invisible.}
\label{fig:comp_sim}
\end{figure*}

We analyze the benefit offered by the refinement stage of CT-Bound. As shown in Fig.~\ref{fig:cnn_trans_ex}, the refinement stage attenuates noisy and inconsistent boundary estimations and strengthens real boundaries compared to the boundary map from the initialization stage. It also makes the color map appear smoother and sharper at color boundaries. The quantitative analysis is shown in Tab.~\ref{tab:diff_cnn}. It draws the same conclusion: the refinement stage increases the ODS F1-score of the boundary map compared to the initialization stage. 

\vspace{-0.05in}
\subsection{Analysis on synthetic and real images}
\label{sec:experiment:synthetic}

We compare the proposed method with the iterative FoJ solver~\cite{verbin2021field}, the traditional edge detector Canny~\cite{canny1986computational}, and other learning-based models, including Boundary Attention~\cite{mia2023boundaries}, PEdger~\cite{fu2023practical}, UAED~\cite{zhou2023treasure}, Restormer~\cite{zamir2022restormer}$\rightarrow$HED~\cite{xie2015holistically}, Restormer~\cite{zamir2022restormer}$\rightarrow$Canny~\cite{canny1986computational}, EDTER~\cite{pu2022edter}, PiDiNet~\cite{su2021pixel}, and HED~\cite{xie2015holistically}. Tab.~\ref{tab:acuracy} shows the quantitative comparison of these methods. Note that our model is trained using images from MS COCO dataset~\cite{lin2014microsoft} with random photon level $\alpha_{\text{train}} \in [2,10]$ and evaluated using images from BSDS500~\cite{arbelaez2010contour} and NYUDv2~\cite{silberman2012indoor} datasets on a specific photon level $\alpha_{\text{test}}$. The proposed approach achieves the highest or near-highest ODS F1-score when the noise level is very high, i.e. $\alpha=2$. Fig.~\ref{fig:comp_sim} shows sample boundary maps estimated from noisy images synthesized from BSDS500~\cite{arbelaez2010contour} and NYUDv2~\cite{silberman2012indoor} datasets. Both results indicate the robustness and generalizability of the proposed method to different image datasets and noise levels. 

\begin{figure}[tb]
\centering
\includegraphics[width=1.0\columnwidth]{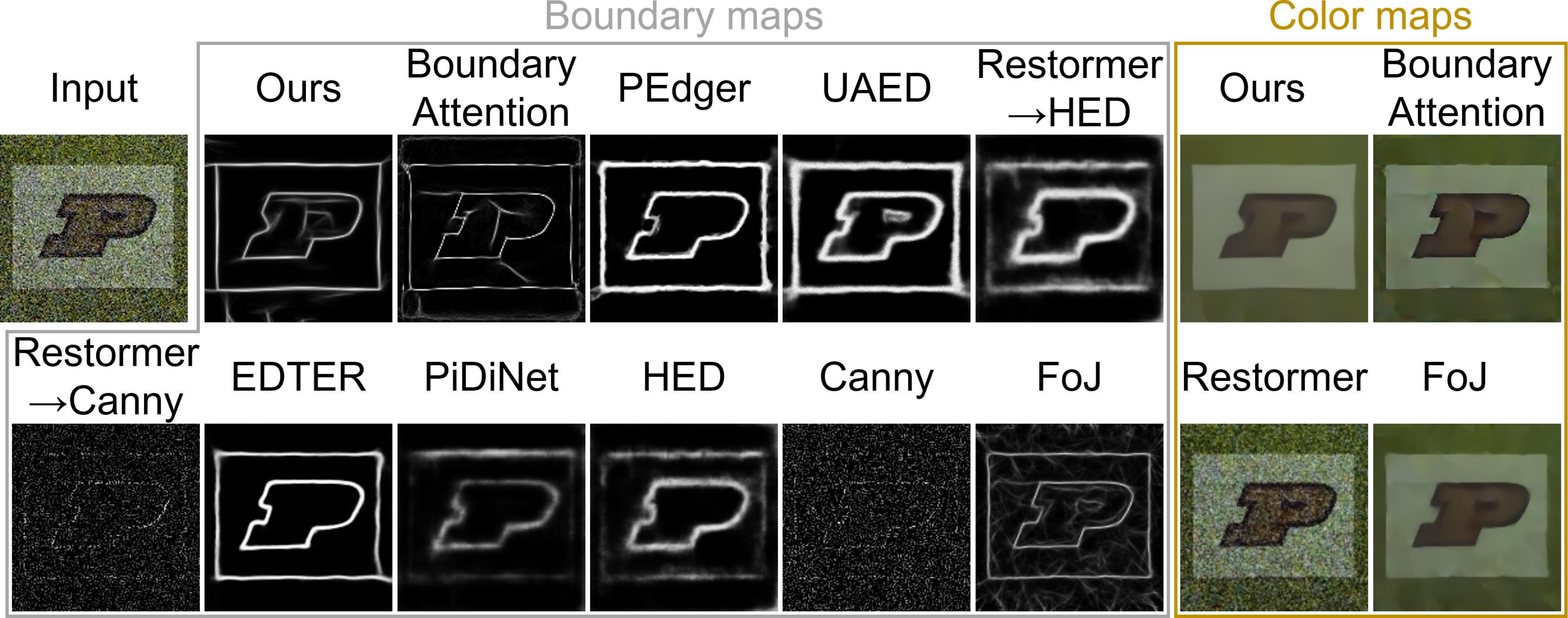}
\caption{Comparison using real captured images. The image is taken at the exposure of 1/10900s and ISO 40.}
\label{fig:real}
\vspace{-0.1in}
\end{figure}

Fig.~\ref{fig:real} shows the boundary maps and color maps estimated from a real image captured by an iPhone 13 Mini camera with a high shutter speed. The boundary map and color map generated from the proposed method have high visual quality without any fine-tuning to the real images. We also upload a video of boundary maps and color maps of a real captured video clip generated by CT-Bound to the URL listed in Sec.~\ref{sec:intro}. 

\vspace{-0.05in}
\section{Conclusion and Limitation}
\label{sec:conclusion}

In this paper, we propose a two-stage boundary detector, CT-Bound, which is a hybrid neural network architecture aiming to achieve robust and accurate boundary detection on extremely noisy images in a single shot. Compared to a variety of models, our method demonstrates the highest or near-highest boundary detection accuracy on benchmark datasets, producing visually clean and crisp boundaries. A limitation we observe is that, when processing videos, there are abrupt changes in detected boundaries across frames. This is because CT-Bound only uses a single frame for processing. The problem can be resolved by introducing temporal consistencies in the model. 

\vspace{-0.05in}
\bibliographystyle{IEEEbib}
\bibliography{refs}

\clearpage

\setcounter{section}{0}
\setcounter{equation}{0}
\setcounter{figure}{0}
\setcounter{table}{0}
\makeatletter
\renewcommand{\thesection}{S\arabic{section}}
\renewcommand{\theequation}{S\arabic{equation}}
\renewcommand{\thefigure}{S\arabic{figure}}
\renewcommand{\thetable}{S\arabic{table}}

\onecolumn

\begin{center}
\textbf{\large Supplemental Materials for CT-Bound: Robust Boundary Detection From Noisy Images Via Hybrid Convolution and Transformer Neural Networks}
\end{center}

In this \textbf{supplementary document}, we first show some samples of our training data in Sec.~\ref{sec:s:data}. Then we report a quantitative comparison for color maps on BSDS500~\cite{arbelaez2010contour} and NYUDv2~\cite{silberman2012indoor} datasets in Sec.~\ref{sec:s:colormapmetric}. Finally we present more qualitative results from BSDS500~\cite{arbelaez2010contour} and NYUDv2~\cite{silberman2012indoor} datasets in Sec.~\ref{sec:s:additionalresults}. 

\vspace{-0.05in}
\section{Training Data}
\label{sec:s:data}

\begin{figure}[htb]
\centering
\vspace{-0.05in}
\includegraphics[width=0.8\columnwidth]{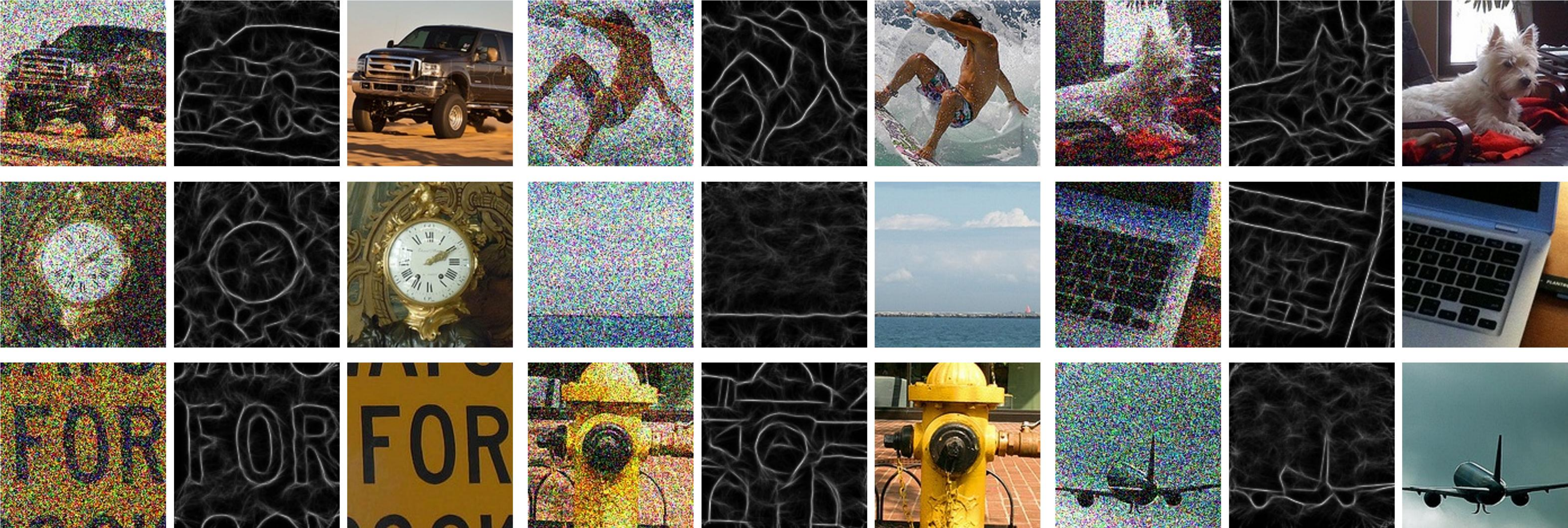}
\vspace{-0.05in}
\caption{Training pair (noisy image, ground truth boundary map, ground truth color map) samples for refinement stage training. The size of each image is $147 \times 147$. The photon level for each image is randomly selected in the range of $[2,10]$.}
\label{fig:s:ref_data}
\vspace{-0.05in}
\end{figure}

We use different datasets for the initialization and refinement stages. For the former stage, our model needs patches as inputs only. We generate noisy patches based on those randomly sampled from FoJ synthetic datasets~\cite{verbin2021field}, which contain basic shapes in grayscale. We assign RGB colors randomly and add noise according to \eqref{eq:noise}. For the latter stage, we use noisy images generated from MS COCO~\cite{lin2014microsoft} with the same noise model. The ground truth boundary maps are generated by running FoJ~\cite{verbin2021field} with the ground truth color maps. Some training pair samples are shown in Fig.~\ref{fig:s:ref_data}. 

\vspace{-0.05in}
\section{Quantitative Results of Color Maps}
\label{sec:s:colormapmetric}

Since different photon levels cause different distribution parameters of pixel values, we normalize the pixel values before calculating the metrics for color maps. Specifically, a color map is normalized through:
\begin{align}
    P^{\prime}(x,y) = \frac{P(x,y)}{\alpha},
\end{align}
where $P(x,y)$ and $\alpha$ is from \eqref{eq:noise}. Then the structural similarity index measure (SSIM), peak signal-to-noise ratio (PSNR), and mean squared error (MSE) are calculated between $P^{\prime}(x,y)$ and $P^*(x,y)$. The quantitative comparison is shown in Tab.~\ref{tab:s:quant_colormap}. 

\begin{table}[htb]
\vspace{-0.05in}
\caption{Quantitative comparison of color maps on noisy images synthesized from BSDS500~\cite{arbelaez2010contour} and NYUDv2~\cite{silberman2012indoor} datasets.}
\label{tab:s:quant_colormap}
\vspace{-0.05in}
\centering
\begin{tabular}{c|c|ccc|ccc}
\hline
\multirow{2}{*}{\shortstack{Photon level\\$\alpha_{\text{test}}$}} & \multirow{2}{*}{Model} & \multicolumn{3}{|c|}{BSDS500} & \multicolumn{3}{|c}{NYUDv2} \\
& & SSIM$\uparrow$ & PSNR(dB)$\uparrow$ & MSE($\times 10^{-2}$)$\downarrow$ & SSIM$\uparrow$ & PSNR(dB)$\uparrow$ & MSE($\times 10^{-2}$)$\downarrow$ \\
\hline
\multirow{4}{*}{2} & FoJ~\cite{verbin2021field} & \underline{\textbf{0.338}} & 10.712 & \textbf{8.626} & \textbf{0.433} & 10.657 & \textbf{8.954} \\
& Restormer~\cite{zamir2022restormer} & 0.152 & 8.492 & 14.624 & 0.157 & 8.567 & 14.312 \\
& Boundary Attention~\cite{mia2023boundaries} & 0.301 & \underline{\textbf{10.952}} & 8.982 & 0.355 & \underline{\textbf{10.791}} & 9.284 \\
& \textbf{Ours} & \textbf{0.332} & \textbf{10.724} & \underline{\textbf{8.606}} & \underline{\textbf{0.467}} & \textbf{10.709} & \underline{\textbf{8.870}} \\
\hline
\multirow{4}{*}{4} & FoJ~\cite{verbin2021field} & \textbf{0.441} & \textbf{17.212} & \underline{\textbf{2.012}} & \textbf{0.572} & 17.619 & \textbf{1.913} \\
& Restormer~\cite{zamir2022restormer} & 0.268 & 14.918 & 3.431 & 0.222 & 14.920 & 3.478 \\
& Boundary Attention~\cite{mia2023boundaries} & \underline{\textbf{0.452}} & \underline{\textbf{17.785}} & 2.031 & 0.560 & \underline{\textbf{18.103}} & 1.926 \\
& \textbf{Ours} & 0.429 & 17.198 & \textbf{2.024} & \underline{\textbf{0.587}} & \textbf{17.740} & \underline{\textbf{1.885}} \\
\hline
\multirow{4}{*}{6} & FoJ~\cite{verbin2021field} & \textbf{0.484} & \textbf{19.780} & \textbf{1.176} & 0.632 & 20.799 & 0.951 \\
& Restormer~\cite{zamir2022restormer} & 0.357 & 17.496 & 2.042 & 0.289 & 17.800 & 1.903 \\
& Boundary Attention~\cite{mia2023boundaries} & \underline{\textbf{0.510}} & \underline{\textbf{20.542}} & \underline{\textbf{1.130}} & \underline{\textbf{0.641}} & \underline{\textbf{21.633}} & \underline{\textbf{0.908}} \\
& \textbf{Ours} & 0.468 & 19.688 & 1.209 & \textbf{0.637} & \textbf{20.918} & \textbf{0.947} \\
\hline
\multirow{4}{*}{8} & FoJ~\cite{verbin2021field} & \textbf{0.507} & \textbf{20.900} & \textbf{0.951} & \textbf{0.667} & 22.427 & \textbf{0.669} \\
& Restormer~\cite{zamir2022restormer} & 0.430 & 19.258 & 1.406 & 0.353 & 19.706 & 1.289 \\
& Boundary Attention~\cite{mia2023boundaries} & \underline{\textbf{0.541}} & \underline{\textbf{21.768}} & \underline{\textbf{0.872}} & \underline{\textbf{0.684}} & \underline{\textbf{23.500}} & \underline{\textbf{0.604}} \\
& \textbf{Ours} & 0.488 & 20.726 & 0.996 & 0.666 & \textbf{22.473} & 0.679 \\
\hline
\end{tabular}
\vspace{-0.05in}
\end{table}

\vspace{-0.05in}
\section{Additional Qualitative Results}
\label{sec:s:additionalresults}

\begin{figure}[htb]
\centering
\vspace{-0.1in}
\includegraphics[width=0.891\columnwidth]{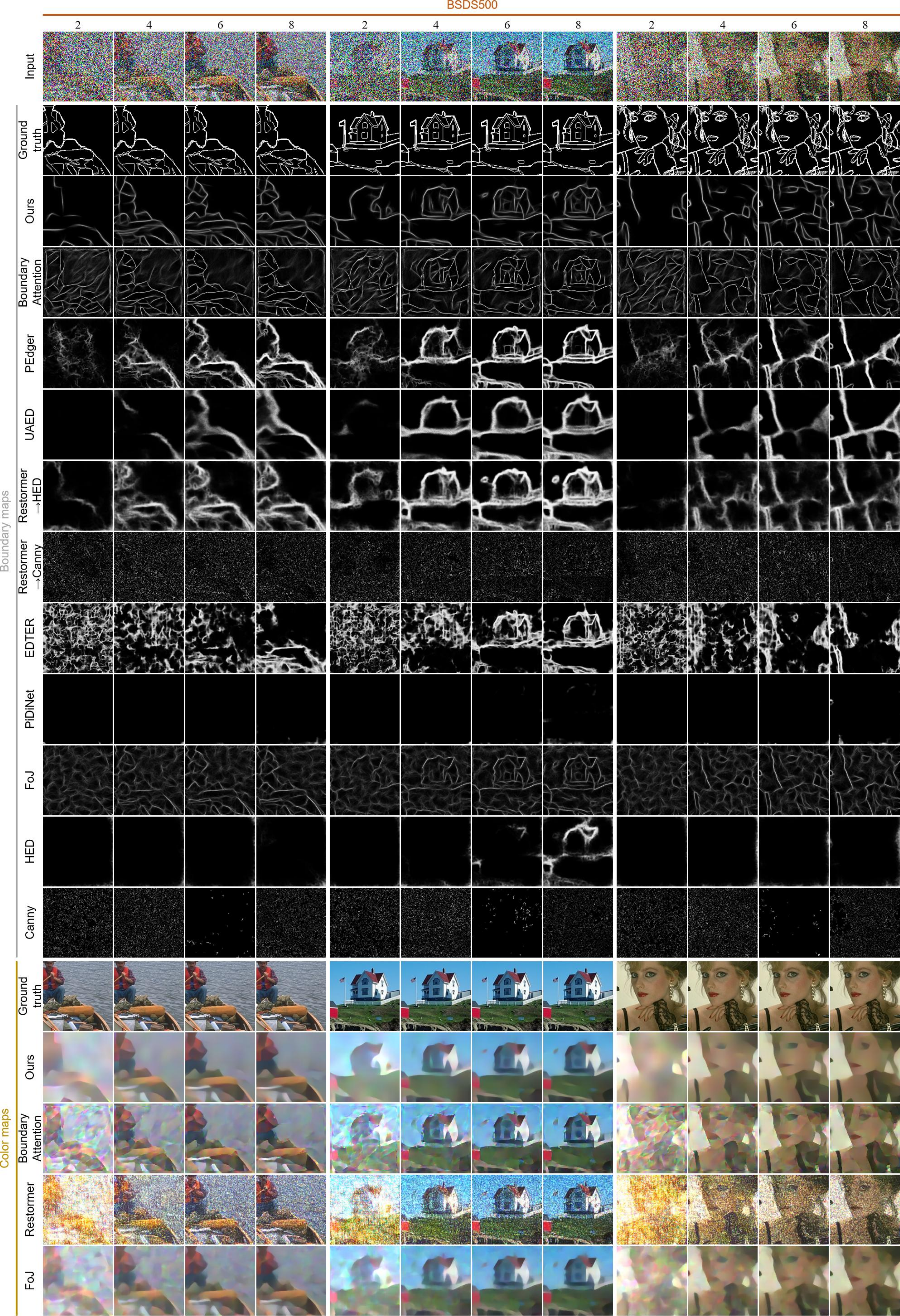}
\vspace{-0.13in}
\caption{Additional qualitative comparison of noisy images synthesized from BSDS500~\cite{arbelaez2010contour} dataset.}
\label{fig:s:comparison1}
\vspace{-0.1in}
\end{figure}

\begin{figure}[htb]
\centering
\vspace{-0.1in}
\includegraphics[width=0.891\columnwidth]{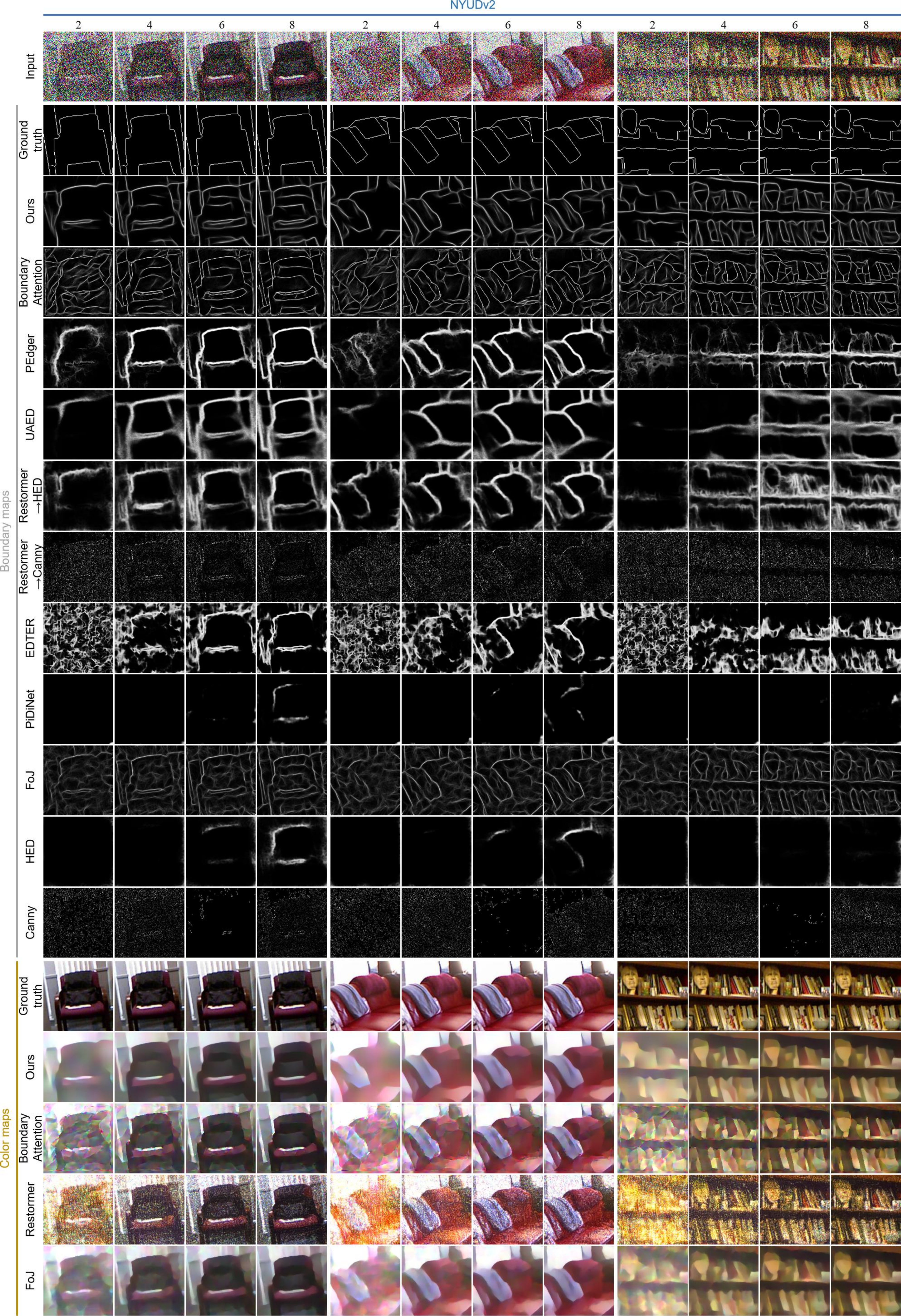}
\vspace{-0.14in}
\caption{Additional qualitative comparison of noisy images synthesized from NYUDv2~\cite{silberman2012indoor} dataset.}
\label{fig:s:comparison2}
\vspace{-0.1in}
\end{figure}

In this section, we show more qualitative comparison results in various photon levels $\alpha_{\text{test}}$, including color maps, in Fig.~\ref{fig:s:comparison1} and Fig.~\ref{fig:s:comparison2}. The proposed method produces more crisp boundaries than some other models, which even have higher ODS F1-scores in Tab.~\ref{tab:acuracy}. 

\end{document}